\documentclass[10pt, a4paper]{article}
\usepackage{lrec2022} % this is the new LREC2022 Style
\usepackage{multibib}

\usepackage{graphicx}
\usepackage{tabularx}
\usepackage{soul}
\usepackage{appendix}
\usepackage[T1]{fontenc}
\usepackage{titlesec}
\titleformat{\section}{\normalfont\large\bfseries\center}{\thesection.}{1em}{}
\titleformat{\subsection}{\normalfont\SmallTitleFont\bfseries\raggedright}{\thesubsection.}{1em}{}
\titleformat{\subsubsection}{\normalfont\normalsize\bfseries\raggedright}{\thesubsubsection.}{1em}{}
\renewcommand\thesection{\arabic{section}}
\renewcommand\thesubsection{\thesection.\arabic{subsection}}
\renewcommand\thesubsubsection{\thesubsection.\arabic{subsubsection}}
\usepackage{epstopdf}
\usepackage[utf8]{inputenc}

\usepackage{hyperref}
\usepackage{xstring}

\usepackage[usenames,dvipsnames]{color}% Enumerate
\usepackage{enumitem}

\usepackage{makecell}
\usepackage{amssymb}% http://ctan.org/pkg/amssymb
\usepackage{pifont}% http://ctan.org/pkg/pifont
\newcommand{\cmark}{\ding{51}}%
\newcommand{\xmark}{\ding{55}}%

\definecolor{babypink}{rgb}{0.96, 0.76, 0.76}
\definecolor{lightgreen}{RGB}{177, 221, 175}
\definecolor{lightgray}{RGB}{223, 223, 223}

\title{The Causal News Corpus: Annotating Causal Relations in Event Sentences from News}

\name{\begin{tabular}{c}
Fiona Anting Tan$^1$, 
Ali Hürriyetoğlu$^2$, 
Tommaso Caselli$^3$,
Nelleke Oostdijk$^4$,\\
Tadashi Nomoto$^5$,
Hansi Hettiarachchi$^6$,
Iqra Ameer$^7$,
Onur Uca$^8$,\\
Farhana Ferdousi Liza$^9$,
Tiancheng Hu$^{10}$
\end{tabular}} 

\address{
$^1$ Institute of Data Science, National University of Singapore, Singapore,
$^2$ Koc University, Turkey,\\
$^3$ Rijksuniversiteit Groningen, Netherlands,
$^4$ Radboud University, Netherlands, \\ 
$^5$ National Institute of Japanese Literature, Japan,
$^6$ Birmingham City University, United Kingdom,\\
$^7$ Centro de Investigación en Computación, Instituto Politécnico Nacional, Mexico, \\
$^8$ Department of Sociology, Mersin University, Turkey, 
$^9$ University of East Anglia, United Kingdom,\\
$^{10}$ ETH Z{\"u}rich, Switzerland
\\
tan.f@u.nus.edu, 
ahurriyetoglu@ku.edu.tr, 
t.caselli@rug.nl,
nelleke.oostdijk@ru.nl,\\
nomoto@acm.org,
hansi.hettiarachchi@mail.bcu.ac.uk,
iqra@nlp.cic.ipn.mx,
onuruca@mersin.edu.tr,\\
f.liza@uea.ac.uk,
tianhu@ethz.ch
}

\abstract{
Despite the importance of understanding causality, corpora addressing causal relations are limited. There is a discrepancy between existing annotation guidelines of event causality and conventional causality corpora that focus more on linguistics. Many guidelines restrict themselves to include only explicit relations or clause-based arguments. Therefore, we propose an annotation schema for event causality that addresses these concerns. We annotated 3,559 event sentences from protest event news with labels on whether it contains causal relations or not. Our corpus is known as the Causal News Corpus (CNC). A neural network built upon a state-of-the-art pre-trained language model performed well with 81.20\% F1 score on test set, and 83.46\% in 5-folds cross-validation. CNC is transferable across two external corpora: CausalTimeBank (CTB) and Penn Discourse Treebank (PDTB). Leveraging each of these external datasets for training, we achieved up to approximately 64\% F1 on the CNC test set without additional fine-tuning. CNC also served as an effective training and pre-training dataset for the two external corpora. Lastly, we demonstrate the difficulty of our task to the layman in a crowd-sourced annotation exercise. Our annotated corpus is publicly available, providing a valuable resource for causal text mining researchers.
 \\ \newline  }

\begin{document}

\maketitleabstract

\section{Introduction}
\label{sec:introduction}

 Causality is a core cognitive concept and appears in many natural language processing (NLP) works that aim to tackle inference and understanding \cite{DBLP:journals/tacl/JoBRH21,dunietz-etal-2020-test,feder2021causal}. Generally, a causal relation is a semantic relationship between two arguments known as cause and effect, in which the occurrence of one (cause argument) causes the occurrence of the other (effect argument) \cite{DBLP:journals/rcs/BarikMO16}. 
 
 Figure \ref{fig:examples} depicts examples of sentences expressing causality, and ones that do not. Notice that causality can be expressed in various ways: The first causal example is signaled by the explicit causal marker \emph{``due to''} while the second example has causality indicated by alternative lexicalizations such as \emph{``created''}. In the last example, an implicit causal relation exists between the verbless clause and the matrix clause, conveying that the dissatisfaction with the package leads to the workers' sit-in. For sentences without causality, they must be missing either a cause or effect argument, or both.

\begin{figure}[!h]
  \centering    
  \includegraphics[scale=0.24]{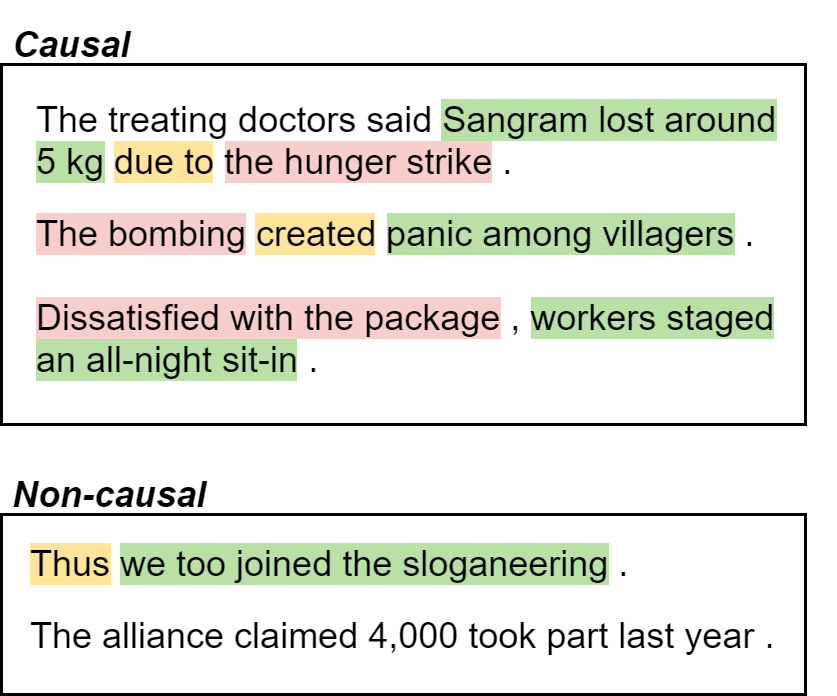}
  \caption{Annotated examples from Causal News Corpus. Causes are in pink, Effects in green and Signals in yellow. Note that both Cause and Effect spans must be present within one and the same sentence for us to mark it as \emph{Causal}.}
\end{figure}\label{fig:examples}

% We believe it is beneficial to create automated solutions that can identify if a particular sentence contains a causal event.

Causal Question Answering and Generation applications \cite{dalal-etal-2021-enhancing,ijcai2019-0695,stasaski-etal-2021-automatically} are some immediate downstream Natural Language Understanding (NLU) applications popular in NLP today. Despite the importance of identifying causality in text, datasets are limited \cite{asghar2016automatic,xu-etal-2020-review,tan-etal-2021-causal,DBLP:journals/corr/abs-2101-06426} and oftentimes, different researchers craft their datasets with different rules, leaving users with no proper way to compare models across datasets.
 
Our work is directed at annotating parts of the multilingual protest news detection dataset \cite{hurriyetoglu-etal-2021-multilingual,hurriyetoglu-etal-2021-challenges,case-2021-challenges} for event causality. Our contributions\footnote{Our corpus and model scripts are available online at \url{https://github.com/tanfiona/CausalNewsCorpus}.} are as follows:
\begin{itemize}
    \item We created the Causal News Corpus (CNC), which is a corpus of event sentences annotated with binary labels indicating whether each sentence contains a causal relationship or not\footnote{Annotation of the corpus is an ongoing effort: We also aim to add information like cause-effect-signal span markings and causal types, and these additional information will also be made public once available.}.
    \item We showed that a neural network built on a state-of-the-art pre-trained language model could predict the causality labels with 81.20\% F1, 77.81\% Accuracy, and 54.52\% Matthews Correlation Coefficient on test set.
    \item We crafted annotation guidelines that align with existing event causality and linguistic schemes. Our experiments show that our dataset is indeed compatible with other causality corpora.
    \item We observed that the layman struggles to replicate the annotations by our experts in a crowd-sourced annotation exercise, demonstrating the difficulty of our task.
\end{itemize}
The rest of the paper is organized as follows: Section \ref{sec:related} reviews the literature on event causality. Section \ref{sec:annotation} describes the compilation and annotation of the corpus. Section \ref{sec:experiments} describes the baselines we created for causal event sentence classification and the experiments that were conducted on CNC and two other datasets. Section \ref{sec:mturk} outlines experiences from a crowd-sourcing annotation exercise. Finally, Section \ref{sec:conclusion} concludes the paper and discusses potential avenues for future work.

\section{Related Work}
\label{sec:related}

The extraction of causality from text remains a significant challenge because semantic understanding of the context and world knowledge is oftentimes needed. In computational linguistics, a lot of effort has been put into extracting causal knowledge from text automatically \cite{blanco2008causal,do2011minimally,kontos1991acquisition,riaz2013toward}. 

Previous work has focused on event causality, creating corpora like CausalTimeBank (CTB) \cite{mirza-etal-2014-annotating} from news, CaTeRS \cite{mostafazadeh-etal-2016-caters} from short stories and EventStoryLine \cite{caselli-vossen-2017-event} from online news articles. Such corpora include rich event features like event class attributes and temporal expressions in TimeML annotation format. However, these corpora are typically limited in size. For example, EventStoryLine (ESL) has 1,770 causal event pairs, CaTeRS has 488 causal links, while CTB has 318 causal event pairs. Our work contributes to this strand of literature by providing more data: We have identified 1,957 causal event sentences, each of which can contain multiple event pairs.

There is also a discrepancy between such event causality corpora and other causality corpora. Penn Discourse Treebank (PDTB) \cite{DBLP:conf/lrec/PrasadDLMRJW08,webber2019penn,prasad2006penn} is a corpus that annotates semantic relations (including causal relations) between clauses, expressed either explicitly or implicitly. This corpus is large (e.g. PDTB-3 has over $7,000$ examples for the \textsc{Contingency.Cause} sense alone) and is potentially useful for training an accurate event sentence classifier. Therefore, we believe that it will be beneficial to align the annotation guidelines of these two corpora types. We did so by constructing event causality annotation guidelines using linguistic rules adapted from PDTB-3. However, our corpus differs from PDTB-3 by focusing on event sentences and accepting more fine-grained arguments that do not necessarily form a clause. This approach of including more varied constructions of causality is similar to the work by BECauSE 2.0 \cite{dunietz-etal-2017-corpus}.

When discussing causality in text, many corpora only focus on explicit relations \cite{DBLP:conf/flairs/GirjuM02,dunietz-etal-2017-corpus}. An explicit causal relation refers to an example where the cause-and-effect relation is expressed by means of a clear connective or some other causal signal. An example was presented earlier in the first sentence of Figure \ref{fig:examples}. Rule-based approaches \cite{DBLP:conf/flairs/GirjuM02,khoo-etal-2000-extracting,DBLP:conf/pakm/SakajiSM08} have been shown to work well on such constructions. For example, researchers have used such approaches to semi-automatically curate datasets for cause-effect related tasks \cite{DBLP:conf/skg/CaoSZ16,DBLP:conf/cikm/HeindorfSWNP20,stasaski-etal-2021-automatically}. However, in the real world, implicit relations are more common but more challenging to identify \cite{hidey-mckeown-2016-identifying}. Therefore, in our work, we do not limit ourselves to a pre-defined list of connectives but rather, include causal examples in more varied linguistic constructions for more realistic use cases. With respect to this, we differ from CTB, which only annotated explicit causal examples.

\section{Compilation \& Annotation}
\label{sec:annotation}

\subsection{Data Source}
CNC builds on the datasets featured in a series of workshops directed at mining socio-political events from news articles: Automated Extraction of Socio-political Events from News (AESPEN) in 2020 \cite{hurriyetoglu-etal-2020-automated,aespen-2020-automated} and Challenges and Applications of Automated Extraction of Socio-political Events from Text (CASE) in 2021 \cite{hurriyetoglu-etal-2021-multilingual,hurriyetoglu-etal-2021-challenges,case-2021-challenges}. The data is based on randomly sampled articles from multiple sources and periods, and all annotations were performed by two annotators, adjudicated and spot-checked by a supervisor, and corrected further semi-automatically~\cite{Hurriyetoglu-et-al-2021-cross-context,Yoruk-et-al-2021-random-sampling}. In total, 869 news documents and 3,559 English sentences were available. We annotated all 3,559 sentences in this paper. We utilized the dataset from CASE 2021 directly and some sentences were not well processed. We noticed that a few examples were actually comprised of more than one sentence and titles were still included in the texts. In our next data release, we will redo the parsing of these examples. 

\subsection{Guidelines}
\label{ssec:guide}

CNC includes annotations for binary causal event classification in sentences; We labeled sentences to be \emph{Causal} or \emph{Non-causal}. Sentences must contain at least a pair of events, defined as as things that happen or occur, or states that are valid, following TimeML~\cite{sauri2006timeml}. For causality, we utilized the definition of \textsc{Contingency} from PDTB-3, which assigns this label for samples where ``one argument provides the reason, explanation or justification for the situation described by the other'' \cite{webber2019penn}. The following four senses were recognized as \emph{Causal}:

\begin{itemize}
    \item \textsc{Cause}: Arg1 and Arg2 are causally influenced but not in a conditional relation. Example connectives are \emph{``because''} and \emph{``since''}. 
    \item \textsc{Purpose}: One Arg describes the goal while the other Arg describes the action undertaken to achieve the goal. Example connectives are \emph{``in order to''} and \emph{``so that''}.
    \item \textsc{Condition}: One Arg presents an unrealized situation, that when realized, would lead to the situation described by the other Arg. Example connectives are \emph{``if''} and \emph{``as long as''}.
    \item \textsc{Negative-Condition}: One Arg presents an unrealized situation, that if it does not occur, would lead to the situation described by the other Arg. Example connectives are \emph{``till''} and \emph{``unless''}.
\end{itemize} 

Since our work is event-based, the senses that provided reasons for the speaker to utter a speech, or the hearer to have a belief (i.e. \textsc{+SpeechAct} or \textsc{+Belief} types) were treated as \emph{Non-causal}.

A major difference between our approach and PDTB is that we permitted non-clausal elements such as phrases in our annotation scheme as arguments as long as these spans meet our definition of event. PDTB is a discourse bank that restricts its argument spans to discourse relations. As noted by \newcite{dunietz-etal-2017-corpus}, causal spans of multiple linguistic forms are excluded from PDTB. For instance, in PDTB, verb signals like \emph{``caused''} are not annotated in the sentence \emph{``Speculation about Coniston has caused the stock to rebound from a low of \$145.''}. Consequently, PDTB also did not annotate the spans before and after \emph{``caused''} as Cause and Effect arguments. Given our focus on news events, more expressions of causality are required for proper interpretation. Therefore, on top of PDTB-approved clauses, we allowed for the following to be recognized as an argument: noun phrases (including nominalizations)\footnote{Some examples of noun phrases describing event arguments: \emph{``the fire''} and \emph{``the protest in May''}.}, and verb phrases (including ones that fall outside of a coordinated structure, provided that the corresponding span of its relation is a noun phrase that would complete the verb phrase to form a clause). In general, these relaxations of rules allowed us to annotate more fine-grained arguments, including ones within clauses.

\begin{table*}[!th]
\centering
%\setlength{\tabcolsep}{1.5pt}
%\begin{tabular}{l|p{5mm}p{10mm}p{18mm}p{18mm}p{3mm}p{8mm}}
\begin{tabular}{lcccccl}
\hline
\multicolumn{1}{l}{Sentence} &  \multicolumn{5}{c}{Causality Tests} & \multicolumn{1}{l}{Label} \\ \cline{2-6}
& Why? & \makecell[l]{Temporal \\ Order} & Counterfact. & \makecell[l]{Ontological \\ Asymmetry} & Linguistic &  \\ \hline 
\makecell[l]{\colorbox{babypink}{The protests spread to 15}  \\ \colorbox{babypink}{other towns} and resulted \\ in \colorbox{lightgreen}{two deaths and} \\ \colorbox{lightgreen}{the destruction of property}.} & \cmark & \cmark & \cmark & \cmark & \cmark & \emph{Causal} \\ \hline
\makecell[l]{\colorbox{lightgray}{Chale was allegedly chased} \\ \colorbox{lightgray}{by a group of about 30} \\ \colorbox{lightgray}{people} and \colorbox{lightgreen}{was hacked to death} \\ \colorbox{lightgreen}{with pangas, axes and spears}.} & \xmark & \cmark & \xmark & \cmark & \xmark & \makecell[l]{\emph{Not} \\ \emph{Causal}}  \\ \hline
\makecell[l]{\colorbox{lightgreen}{The strike will continue} \\ till \colorbox{babypink}{our demands are conceded}.} & \cmark & \cmark & \cmark & \cmark & \cmark & \makecell[l]{\emph{Causal} \\ (Neg. \\ Cond.)} \\

\hline
\end{tabular}
\caption{Examples illustrating the applications of the Tests for Causality (Grivaz, 2010; Dunietz et al., 2017). Cause in pink, Effect in green; potential Cause in gray. Signals are not marked.}\label{tab:corpus_layers}
\label{tab:exe}

\end{table*}

To ground our annotations, we utilized the five tests for causality based on the work by \newcite{grivaz-2010-human} and \newcite{dunietz-etal-2017-corpus}. In Table~\ref{tab:exe}, we report some examples to illustrate the applications of the causality tests and the resulting classification labels. Additional details and examples are provided in Appendix~\ref{adx:annotated-examples}.%, we demonstrate the usage of the framework to justify the expected classification label for Examples \ref{eg_1} to \ref{eg_8}.
% These tests serve as useful checks for cause-effect span annotation in the subsequent sections too.

\paragraph{Five Tests for Causality}
\begin{enumerate}
    \item \textbf{Why:} The example is not causal if the reader is unable to construct a ``Why'' question regarding the Effect.
    \item \textbf{Temporal order:} The example is not causal if the Cause does not precede the Effect in time.
    \item \textbf{Counterfactual:} The example is not causal if the Effect is equally likely to occur or not occur without the Cause.
    \item \textbf{Ontological asymmetry:} The example is not causal if the reader can readily swap the Cause and Effect claims in place.
    \item \textbf{Linguistic:} The example is likely to be causal if it can be rephrased into ``X causes Y'' or ``Due to X, Y.''
\end{enumerate}

\paragraph{\textsc{Cause} types} Quoting the PDTB-3 annotation manual \cite{webber2019penn}, ``\textsc{Cause.Reason} (\textsc{Cause.Result}) is used when Arg2 (Arg1) gives the reason, explanation or justification, while Arg1 (Arg2) gives its effect''. Within the \textsc{Cause} sense, we noticed that its sub-senses, \textsc{Cause.Reason} and \textsc{Cause.Result}, fit into the five tests framework well. In particular, we were able to construct a question like ``Why did \texttt{<effect>}?'', and answer with ``Because (of) \texttt{<cause>}.'' in an extractive manner. 

However, we required some relaxations in the construction of the ``Why'' test to be able to include other senses into our corpus. We highlight the relaxations as follows:

\paragraph{\textsc{Purpose} types} We permitted the answer to include additional words like ``In order to (achieve the goal of) \texttt{<cause>}''. In this setting, the cause is a goal, or a justification for the effect action. See Appendix~\ref{adx:annotated-examples} Example \ref{eg_3}.

\paragraph{\textsc{Condition} types} We allowed the relaxation of the question to include modal terms, for example: ``Why could/would \texttt{<effect>} occur?''. See Appendix~\ref{adx:annotated-examples} Example \ref{eg_7}.

\paragraph{\textsc{Cause.NegResult} and \textsc{Negative-Condition} types} Quoting the PDTB-3 annotation manual \cite{webber2019penn}, ``\textsc{Cause.NegResult} is used when Arg1 gives the reason, explanation or
justification that prevents the effect mentioned in Arg2''. For \textsc{Cause.NegResult} and \textsc{Negative-Condition} senses, we allowed a further relaxation of the question to include ``prevent'' or ``cause-to-end'' terms, as in: ``Why was \texttt{<effect>} prevented?'', ``Why could be \texttt{<effect>} prevented?'' or ``Why did \texttt{<effect>} end?''. See Appendix~\ref{adx:annotated-examples} Example \ref{eg_8}.

For samples with multiple sentences, we annotated each sentence independently, that is, we do not consider cross-sentence causality. We acknowledge that by omitting cross-sentence examples, we potentially lose out between a 1/10 to a 1/5 of additional causal examples, based on proportions estimated from CTB and ESL respectively. Titles were also not annotated.

\subsection{Workflow}
\label{ssec:workflow}
\begin{figure}[!h]
\begin{center}
%\fbox{\parbox{6cm}{
%This is a figure with a caption.}}
% old picture \includegraphics[scale=0.5]{lrec2020W-image1.eps} 
\includegraphics[scale=0.135]{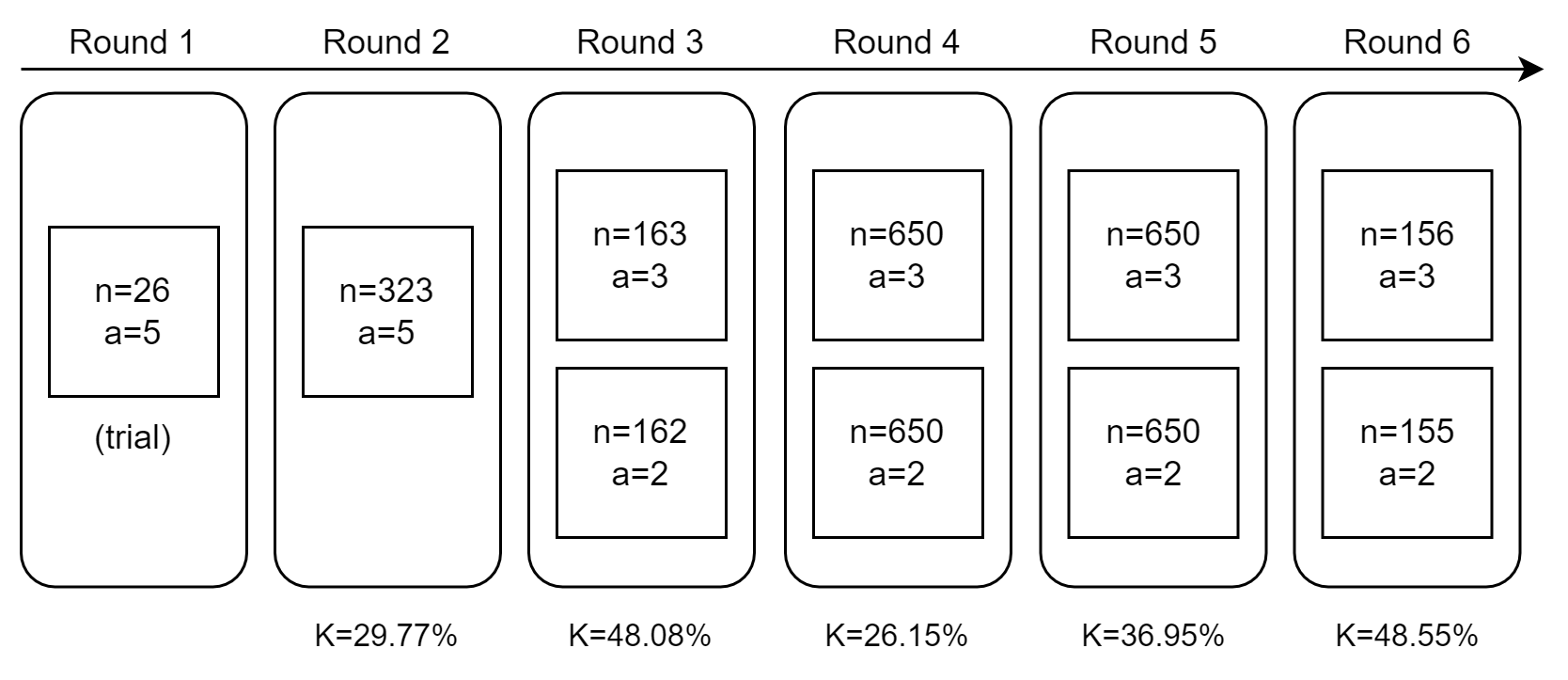} 

\caption{Summary of annotation workflow. $n$: sample size for the subset. $a$: number of annotators for the subset. $\kappa$: Krippendorff's Alpha score for samples per round. Round 1 was a trial training round where annotations were discarded. All subsequent rounds were for the training set except Round 6 which was for the test set.}
\label{fig:summary}
\end{center}
\end{figure}

Five annotators and one curator were involved in the annotation effort. Figure~\ref{fig:summary} summarizes the different annotation iterations to build CNC.

\subsubsection{Initial Training}
The initial training phase for annotators corresponds to Round 1 in Figure \ref{fig:summary}. Guidelines were presented at the beginning of the annotation phase, and examples outside of CNC dataset were ran through with annotators. We found it difficult to arrive at a consensus on the final label across annotators during this initial discussion. This experience motivated us to update our guidelines after each round of annotation to continually improve annotators' understanding of the guidelines.

\subsubsection{Guidelines Refinement}
The difficulty in achieving alignment across annotators, especially at the initial stages, arose from the following three issues: Firstly, since our task challenges literacy, comprehension, and cognitive understanding of a sentence, it is a complex and involved task. Thus, a significant amount of time is required for careful reading of the annotation manual and examples to improve familiarity. Secondly, sensitivity to causal cues and signals (including implicit signals) takes time to develop, and might not be apparent to some until later rounds. Thirdly, different annotators rely on different guidelines or feedback mechanisms to improve their accuracy to identify causal labels faithful to the guidelines. Therefore, we were motivated to update our guidelines in an iterative fashion after each annotation round to refine the descriptions and include more examples that would help improve alignment across annotators. A group discussion always followed at the end of each annotation round for annotators and curator to share feedback and agree on the final guidelines. We also allowed annotators to skip examples if they were unable to come up with a suitable label.

For any major changes in guidelines that might affect previous rounds of annotations, the curator identified such examples and discussed with all annotators before the final annotation is amended. In total, 33 examples had their labels changed in this manner.

\subsubsection{Final Annotation}
In Round 2, we achieved a Krippendorff's Alpha ($\kappa$) score of 29.77\% across five annotators. After the second annotation round, we split the five annotators into two groups to cover more data within our time frame. Concurrently, we also noticed an improvement in the agreement scores to 48.08\%. By the third annotation round, the guidelines had no further changes. Surprisingly, we noticed agreement scores fell to 26.15\% in Round 4. This could be due to the much larger dataset presented to annotators in this round, causing reduced focus. In fact, we noticed the main reason for the drop arose from one annotator in the first group. Excluding his contribution, the agreement increased to a reasonable score of 39.34\%. After working out misunderstandings, the subsequent two rounds of annotations achieved consistent agreement scores of 36.95\% and 48.55\%. Across all rounds, our dataset has an agreement score of 34.99\%.

In our final corpus, every example was annotated by at least two annotators. If only one annotation is present because the other annotator skipped that example, the curator's vote is final. The curator also resolved any ties present throughout the annotation exercise. In the end, we obtained 3,559 annotated sentences, with 1,957 marked as \emph{Causal} and 1,602 marked as \emph{Non-causal}. This is reflected in the first row of Table \ref{tab:datasizes}, alongside data sizes for two other external datasets we experimented with in the next section. We followed the train-test split of our data source, and also provide the breakdown of data sizes by class labels in Table \ref{tab:datasizes}.

\begin{table}[!h]
\begin{center}
\begin{tabular}{p{22mm}llll}
\hline
 & \emph{Causal} & \emph{Non-causal} & Total \\ \hline
CNC & 1,957 & 1,602 & 3,559 \\
CNC Training & 1,781 & 1,467 & 3,248 \\
CNC Test & 176 & 135 & 311 \\
PDTB-3 & 9,917 & 18,358 & 28,275 \\
CTB & 318 & 1,418 & 1,736\\\hline
\end{tabular}
\caption{Data sizes by class label.}
\label{tab:datasizes}
 \end{center}
\end{table}

\section{Experiments}
\label{sec:experiments}
% Causal & Random Baselines: http://localhost:8888/notebooks/61%20Challenges/2022_CASE_/Documentation/Paper/202206_LREC/Summary%20Statistics.ipynb

\begin{table*}[!h]
\begin{center}
\begin{tabular}{lllllll}\\\hline
\# & Training Set & F1 & P & R & Acc & MCC \\\hline
1 & All \emph{Causal} & 72.28 & 56.59 & \textbf{100.00} & 56.59 & 0.00\\
2 & Random & 55.72 & 56.61 & 54.92 & 50.66 & 0.00 \\
3 & CNC Training & \textbf{81.20} & 78.01 & 84.66 & \textbf{77.81} & \textbf{54.52} \\ %outs/test/eval_results.json
4 & PDTB-3 & 55.43 & \textbf{81.32} & 42.05 & 61.74 & 32.09 \\ %outs/pdtb/eval_results.json
5 & PDTB-3 Bal & 64.45 & 77.60 & 55.11 & 65.59 & 34.75 \\
6 & CTB & 27.36 & 80.56 & 16.48 & 50.48 & 17.49 \\ %outs/pdtb/eval_results.json
7 & CTB Bal & 64.05 & 75.38 & 55.68 & 64.63 & 32.13\\ %outs/ctb_r/eval_results.json
\hline
\end{tabular}
\caption{Metrics from predictions on CNC Test Set using various settings. F1, Precision (P), Recall (R), Accuracy (Acc) and Matthews Correlation Coefficient (MCC) are reported in percentages (\%). Highest score per column is indicated in boldface.}
\label{tab:results}
 \end{center}
\end{table*}

In this section, we present a battery of baseline models and results of causal sentence classification on the CNC data and the two existing news corpora, PDTB-3 \cite{webber2019penn} and CTB \cite{mirza-etal-2014-annotating}.

\subsection{Evaluation Metrics}
\label{sec:evaluation}

For all experiments, we evaluated the predictions against true labels using five evaluation metrics: Precision (P), Recall (R), F1, Accuracy (Acc) and Matthews Correlation Coefficient (MCC) \cite{matthews1975comparison}. Formulas are provided below in Equations (1) to (5), where $TP$/ $TN$/ $FP$/ $FN$ stands for True Positive/True Negative/False Positive/ False Negative respectively. All metrics were calculated using the \texttt{scikit-learn} \cite{scikit-learn} package.

\begin{equation}
    P = \frac{TP}{TP + FP}
\end{equation}

\begin{equation}
    R = \frac{TP}{TP + FN}
\end{equation}

\begin{equation}
    F1 = \frac{2 \times P \times R}{P+R}
\end{equation}

\begin{equation}
    Acc = \frac{TP + TN}{TP + TN + FP + FN}
\end{equation}

% \resizebox{0.98\columnwidth}{!}{
\begin{equation}
    MCC = \frac{TP\times TN-FP\times FN}{\sqrt{(TP+FP)(TP+FN)(TN+FP)(TN+FN)}}
\end{equation}
% }

\subsection{Models}
\subsubsection{Dummy Baselines}

To create naive baselines, we obtained dummy predictions using two approaches. For the first row indicated by ``All \emph{Causal}'' of Table \ref{tab:results}, we predicted all examples to be \emph{Causal}, which is the majority label of our dataset. Notice that the Recall score is perfect (100\%), because all \emph{Causal} examples will be identified if we predict all as \emph{Causal}. However, Precision is low (56.59\%), because all \emph{Non-causal} examples were wrongly identified as \emph{Causal}. In the subsequent discussions, we focus our key evaluation metrics to be F1, Accuracy and MCC. For the second row, we randomly assigned the \emph{Causal} or \emph{Non-causal} labels to each example based on the ground truth distribution from CNC's training set, averaged over $1,000$ runs.

\subsubsection{Neural Network Baselines}
We constructed a baseline model by fine-tuning the pre-trained Bidirectional Encoder Representations from Transformers (BERT) model \cite{devlin-etal-2019-bert}. In particular, we used \texttt{bert-base-cased} from Huggingface \cite{wolf-etal-2020-transformers} with AdamW Cross Entropy Loss. The learning rate was $5e-05$ with linear scheduling. Train batch size was $32$, and number of epochs was $10$. The model pipeline is as follows: Tokenized sentences aree run through a BERT transformer module. The pooled output is obtained by taking the hidden state corresponding to the first token. Finally, a binary classification head is applied to obtain the predicted logits.

Additionally, we constructed a Long-Short Term Memory (LSTM) model \cite{hochreiter1997long} with fastText embeddings \cite{bojanowski2017enriching}. The LSTM model performed on par with the BERT model for F1 but had lower MCC values than the BERT model. Overall, the BERT model outperformed the LSTM model. Details about the LSTM baseline and its evaluations are in Appendix \ref{adx:other-baseline-models}. Our subsequent analyses focuses on the BERT baseline.

\subsection{Training on CNC}
From the third row in Table \ref{tab:results}, when we trained on our training set and applied on the test set, we achieved a reasonable F1 score of 81.20\%, Accuracy of 77.81\%, and MCC of 54.52\%, each exceeding the dummy baselines scores reflected in the first two rows. This finding demonstrates that our annotations are internally consistent and reliable for a simple baseline model to learn effectively.

%\subsubsection{Training on External Corpora}
\subsection{Training on External Corpora}
\label{sec:exp-ext}

% Code: "Get PDTB Data.ipynb", "Get CTB for CASE.ipynb"
In this subsection, we check the transferability of existing event causality and causality corpora on CNC. To conduct the experiment, we first trained on only the external corpus (i.e. PDTB-3 or CTB) and then applied the trained model to predict causal labels on CNC test set. 

For PDTB-3, we trained on 28,275 examples (9,917 as \emph{Causal} and 18,358 as \emph{Non-causal}). We obtained this number of examples by selecting all causal examples based on the four senses described in Section \ref{ssec:guide}. Next, we added the \emph{Non-causal} examples and performed de-duplication such that if two examples from the same document overlap in some part but have a different causal label, we dropped the \emph{Non-causal} example. Once the model is trained, we used it to predict the causal labels of CNC test set. According to Row 4 of Table \ref{tab:results}, there is some overlap in the annotation schemes of PDTB and CNC. Therefore, the model was able to achieve Accuracy and MCC scores of 61.74\% and 32.09\%, which are higher than the first two dummy baselines. However, the F1 score is only at 55.43\%, comparable to the Random baseline and worse than the All-\emph{Causal} assignment. The low F1 score is driven by the low Recall score, which indicates that a model trained on PDTB-3 is unable to detect most of the true \emph{Causal} examples. However, its high Precision score indicates the model does return high-quality predictions. The low Recall could be explained by the fact that we included guidelines in our annotation schema that are different from PDTB's schema. For example, we were not strict about arguments having to be a proper clause. That is, for a sentence like \emph{``John caused the fire and walked away.''}, we would have marked it as \emph{Causal}. However, since \emph{``John''} and \emph{``the fire''} are not clauses, PDTB would not mark this example as \emph{Causal}. As a result, a model trained on PDTB would also recognize fewer causal constructions.

To tackle the issue of imbalanced class distribution in PDTB-3, which might affect model performance, we reran an experiment with randomly sampled \emph{Non-causal} examples to match the number of \emph{Causal} examples. This corpus is referred to as ``PDTB-3 Bal'' ($n=19,834$) in Row 5 of Table \ref{tab:results}. All metrics, except Precision, improved relative to the imbalanced PDTB-3 setting.

We repeated the experiment with examples obtained from CTB. For this corpus, we had 1,736 examples (318 as \emph{Causal} and 1,418 as \emph{Non-causal}) to train with. For sentences marked with ``CLINK'' in CTB, we denoted them as \emph{Causal}. All remaining sentences were marked as \emph{Non-causal}. Note a possible complication arises because while all \emph{Non-causal} examples are single sentences, \emph{Causal} examples might span across a few sentences. Performance was dismal, returning comparable or worse scores for F1 and Accuracy than both dummy baselines. We hypothesized that this poor outcome is a result of CTB's extremely imbalanced class distribution. Indeed, when we retrained on a balanced sample, ``CTB Bal'' ($n=636$), we observed performance improvements across all metrics in the last row of Table \ref{tab:results}. A reasonable F1 score of 64.05\%, Accuracy of 64.63\% and MCC of 32.13\% was obtained, similar to or slightly better than PDTB's performance, despite CTB Bal being much smaller in size.

To conclude, we trained models on external corpora and used them directly to make predictions on our proposed CNC corpus. This demonstrates the transferability of existing causality corpora on our corpus. Nonetheless, since both external corpora contain multi-sentence examples and differ slightly in annotation guidelines, some performance differences are expected.

\subsection{Training \& Testing on External Corpora}

\begin{table*}[!h]
\begin{center}
\begin{tabular}{llllllllllll}
\hline
Training Set  & \multicolumn{8}{c}{Test Set} \\ 
\cline{2-10} 
 & \multicolumn{4}{c}{F1} & & \multicolumn{4}{c}{MCC} \\
\cline{2-5} \cline{7-10}
 & CNC & PDTB-3 & CTB Bal & TRF$\uparrow$ & & CNC & PDTB-3 & CTB Bal & TRF$\uparrow$ \\ 
\hline
 CNC & \textbf{83.46} & 58.38 & 80.65 & \textbf{74.16} & & \textbf{61.71} & 30.68 & 59.11  & \textbf{50.50}\\
PDTB-3 & 56.45 & \textbf{74.45} & 60.79 & 63.90 & & 35.86 & \textbf{61.36} & 32.49 &{43.24}  \\
CTB Bal & 59.10 & 49.21 & \textbf{83.41} & 63.90 & & 32.10 & 17.48 & \textbf{65.01} & {38.20} \\ \hline
\end{tabular}
\caption{Metrics from predictions using different train and test sets. Diagonals per sub-table (i.e. when training set and test sets are the same) refers to 5-folds CV experiments. F1 and Matthews Correlation Coefficient (MCC) are reported in percentages (\%). Transferability Rate (TRF) indicates how well a model trained on a given corpus works for unseen, external datasets: For example, the TRF for CNC is given by the average of its performance on PDTB-3 and CTB Bal, based on a particular metric (such as F1 or MCC). Highest score per column is indicated in boldface.}
\label{tab:cross_results}
 \end{center}
\end{table*}

We continue to study the compatibility of CNC with the two external corpora. In this section, we study if a model trained on CNC will be helpful for event sentence causality prediction on unseen datasets like PDTB-3 and CTB.

For each of the three corpora, we trained on the whole corpus and applied the trained model on the remaining two corpora. For example, we trained on all examples from CNC (3,559 examples) and used the model to predict labels for PDTB-3 and CTB Bal. We experimented with the balanced corpus of CTB because the results of the  experiments presented in Section~\ref{sec:exp-ext} shown that the full CTB returns extremely poor performance when used for training. We also included the 5-Fold cross-validation (CV) scores as a reference: For each dataset, we randomly split the examples into five validation sets, and for each round, the corresponding remaining samples were used for training.

Table \ref{tab:cross_results} reflects key experimental results. As expected, for every corpus, the CV training setup always returns the best performance. After which, for both PDTB-3 and CTB Bal, we noticed that a model trained on CNC always returns a higher performance across metrics. For example, when testing on CTB Bal, a model trained on PDTB-3 achieved 60.79\% while a model trained on our corpus achieved 80.65\% F1 score. We introduced a metric called Transferability Rate (TRF), which computes the mean score across test sets. TRF indicates how well a model trained on a given corpus works on average for unseen, external datasets. CNC obtained the highest TRF across the three corpora for both F1 and MCC, reflecting good transferability.

Our findings once again demonstrate that there exists some differences in the annotation schemes for causality, which reduces the transferability of trained models between corpora. Nevertheless, our corpus is the most transferable between the event causality (i.e. CTB) and linguistic causality (i.e. PDTB) corpora studied. Therefore, our annotation guidelines and dataset can serve as a bridge between these two strands of causality corpora.

\subsection{Using CNC for Pre-Training}

\begin{table*}[!h]
\centering
% \resizebox{\columnwidth}{!}{
\begin{tabular}{lllllll}\hline
Dataset & PTM & F1 & P & R & Acc & MCC \\\hline
PDTB & \texttt{bert-base-cased} & 74.45 & \textbf{76.76} & 72.31 & 82.60 & 61.36 \\
 & CNC-PTM & \textbf{75.19} & 75.73 & \textbf{74.69} & \textbf{82.71} & \textbf{61.95} \\\hline
CTB Bal & \texttt{bert-base-cased} & 83.41 & 77.25 & \textbf{90.95} & 81.91 & 65.01 \\
 & CNC-PTM & \textbf{84.68} & \textbf{80.14} & 90.02 & \textbf{83.80} & \textbf{68.30}\\\hline
\end{tabular}
% }
\caption{Metrics from pre-trained model (PTM) experiments. F1, Precision (P), Recall (R), Accuracy (Acc) and Matthews Correlation Coefficient (MCC) are reported in percentages (\%).  Highest score per column per dataset is indicated in boldface.}
\label{tab:ptm}
\end{table*}

Since we believe CNC is transferable, we hypothesize that CNC will also be useful for training a pre-trained model (PTM). Later, this PTM can be fine-tuned onto any other causality-based corpora. We term the BERT PTM updated on all examples of CNC as `CNC-PTM'. CNC-PTM was then fine-tuned on each external dataset (i.e. PDTB and CTB Bal). We again used the 5 CV setup\footnote{For each external dataset and each fold, CNC-PTM was fine-tuned on the training set and applied onto the validation set. This means that for each external dataset, we would have 5 fine-tuned models at the end of the CV experiment.}, but reduced the training epochs to 2. We chose this low number in order to keep our model intact. The fine-tuning step only serves to adjust the model's alignment to better fit the new data distribution.

From Table \ref{tab:ptm}, we indeed found improvements in the average performance across folds for both PDTB and CTB Bal. Logically, CNC-PTM is better initialized for causal sentence classification than the general \texttt{bert-base-cased} PTM for two reasons: (1) It had access to more causal examples and, (2) Its structure is more adapted for the classification task. Thus, performance improvements are expected. These findings again support our claim that CNC is generalizable to out-of-distribution datasets. Therefore, it was useful for updating a PTM.

\section{Crowd-Sourcing Annotations}
\label{sec:mturk}
% Code: "Random Sample for MTurk.ipynb", "Review MTurk.ipynb"

\begin{table}[!h]
\centering
\resizebox{\columnwidth}{!}{
\begin{tabular}{llllll}\hline
 & F1      & P & R  & Acc & MCC\\\hline
  All \emph{Causal} & \textbf{66.00} & \textbf{50.00} & \textbf{100.00}  & \textbf{48.40} & -3.89 \\
 Majority & 61.97 & 47.83 & 88.00  & 46.00 & -14.74 \\
Each Vote & 59.31 & 48.96  & 75.20  & 48.40 & \textbf{-3.79} \\\hline
\end{tabular}}
\caption{Metrics from crowd-sourced workers for a subset of 50 examples. F1, Precision (P), Recall (R), Accuracy (Acc) and Matthews Correlation Coefficient (MCC) are reported in percentages (\%).  Highest score per column is indicated in boldface.}
\label{tab:mturk}
\end{table}

We experimented with crowd-sourced workers using Amazon Mechanical Turk (MTurk). We randomly selected 50 examples, of which 25 were \emph{Causal} and 25 were \emph{Non-causal}. For each causal label, we selected 13 examples that were more complex (with the percentage of the vote for the label being $\leq0.75$, while the remaining 12 examples were `easier' examples with high agreements). The annotation manual was provided to the MTurk workers, and no restrictions were set on the type of workers we allowed for. We requested for five unique annotators to be assigned to each example. Therefore, 250 annotations were received\footnote{Although our dataset was balanced, crowd-sourced workers returned a skewed distribution in votes: 192 votes for \emph{Causal} and 58 votes for \emph{Non-causal}.}, of which we had 44 unique workers. 

Kappa scores were low at 1.62\%. Evaluation metrics of these crowd-sourced annotations against our expert labels are reflected in Table \ref{tab:mturk}. The second row reflects predictions taken based on the majority vote (i.e. mode out of five votes per example). The third row reflects outcomes when we compared each of the 250 annotations with our true labels independently. In both cases, performance of crowd-sourced workers is poor. Since most annotators labelled examples as \emph{Causal}, the results are similar to the results from an All-\emph{Causal} dummy baseline (shown in the first row). In fact, in terms of F1, Accuracy, and MCC, the workers' scores are close, if not worse than, the All \emph{Causal} baseline. As highlighted in Section \ref{ssec:workflow}, the task is challenging and requires a thorough understanding of the annotation guideline to do well. Given that MTurk provides a low monitoring environment with no room for iterative feedback, it is understandable that crowd-sourced workers perform poorly (close to random assignment) at their tasks.

Nevertheless, this experiment provided a few learning points: Firstly, our manual annotations are not easy to identify by a layman, and skilled annotators are required. Similar to our annotation experience, it is necessary to have repeated discussions and feedback sessions with annotators so as to improve agreement scores. Secondly, compared to a BERT model trained on internal examples or external examples annotated with similar logic, the layman identifies causality poorly\footnote{In terms beating dummy baselines, especially for Accuracy and MCC metrics.}. \newcite{caselli-inel-2018-crowdsourcing} argues that discrepancies in the quality of crowd-sourced annotations from experts should be used as an estimate on how complex the task is. Together, these findings highlight that our corpus is a unique and valuable resource, requiring an immense amount of time and effort to create with experts.

\section{Conclusion}
\label{sec:conclusion}
Causality is an important cognitive concept that deserves to have a corpus dedicated to the definition and identification. CNC focuses on annotating causality in text, and so far, we have created an annotation schema to identify if event sentences contain causal relations or not. 

CNC's annotation guidelines are constructed based on linguistic rules, and covers a wider array of causal linguistic constructions than previous works. We also demonstrated transferability between CNC and existing datasets that include causal relations like CTB and PDTB. Finally, the binary causal event classification task is challenging: layman workers only achieved close poor performance in our crowd-sourced annotation experiment. Therefore, CNC, which has been annotated by experts, is a valuable resource for researchers working in the causality and event causality space. CNC could potentially be relevant to NLU  researchers working on downstream tasks too.

We are also organizing a shared task using CNC to promote the development of automatic causal text mining solutions\footnote{Our shared task competition page is at \url{https://codalab.lisn.upsaclay.fr/competitions/2299}.}. In our next dataset release, we will clean some parts of our current dataset and perform additional curation of mispredicted examples in CV. We are also in the midst of adding fine-grained annotations, such as cause-effect-signal spans and causal concept labels.

\newpage
\section{Acknowledgements}

This project is supported by the National Research Foundation, Singapore under its Industry Alignment Fund – Pre-positioning (IAF-PP) Funding Initiative. Any opinions, findings and conclusions or recommendations expressed in this material are those of the author(s) and do not reflect the views of National Research Foundation, Singapore.

\section{Bibliographical References}\label{reference}
\label{main:ref}
\bibliographystyle{lrec2022-bib}
\bibliography{bibs/anthology, bibs/custom, bibs/languageresource, bibs/lrec2022-example}

% \section{Language Resource References}
% \label{lr:ref}
% \bibliographystylelanguageresource{lrec2022-bib}
% \bibliographylanguageresource{languageresource}

\newpage
\section*{Appendix}
\begin{appendices}

\section{Annotated Examples} \label{adx:annotated-examples}
The following are some examples of the expected causal label for each sentence, along with justifications using the Five Tests highlighted in Section \ref{ssec:guide}.

\begin{enumerate}[label*=(\arabic*),series=example]
  \item{His attackers allegedly drank his blood .\label{eg_1}}
\end{enumerate}
\begin{itemize}[leftmargin=1cm]
    \item \emph{Non-causal}: There is only one event in this sentence.
\end{itemize}

\begin{enumerate}[label*=(\arabic*),resume*=example]
  \item{The protests spread to 15 other towns and resulted in two deaths and the destruction of property .\label{eg_2}}
\end{enumerate}
\begin{itemize}[leftmargin=1cm]
    \item \emph{Causal}: This sentence contains causal events.
\end{itemize}
\begin{enumerate}[leftmargin=1cm]
    \item \textbf{Why:} Why were there “two deaths and the destruction of property”? Because “the protests spread to 15 other towns”.
    \item \textbf{Temporal order:} Protests must spread (Cause) before deaths and destruction (Effect) can occur.
    \item \textbf{Counterfactual:} Deaths and destruction (Effect) are unlikely to occur if the protests did not spread (NegCause).
    \item \textbf{Ontological asymmetry:} Using Cause as an Effect to construct a question: Why did “The protests spread to 15 other towns”? Because of deaths and destruction -- does not answer the question.
    \item \textbf{Linguistic:} “The protests spread to 15 other towns” causes “two deaths and the destruction of property”.
\end{enumerate}

\begin{enumerate}[label*=(\arabic*),resume*=example]
  \item{The three-member Farlam Commission, chaired by retired judge Ian Farlam, was established by President Jacob Zuma to probe into the violence and the deaths of 44 people in wage-related protests.\label{eg_3}}
\end{enumerate}
\begin{itemize}[leftmargin=1cm]
    \item \emph{Causal}: This sentence contains causal event of the \textsc{Purpose} sense.
\end{itemize}
\begin{enumerate}[leftmargin=1cm]
    \item \textbf{Why:} Why was ``the three-member Farlam Commision ... established by President Jacob Zuma''? In order ``to probe into the violence and the deaths of 44 people in wage-related protests''.
    \item \textbf{Temporal order:} The probe purpose (Cause) must occur before establishing the commission (Effect).
    \item \textbf{Counterfactual:} Establishing the commission (Effect) is unlikely to occur if the President did not have the goal to probe into the protests (NegCause).
    \item \textbf{Ontological asymmetry:} Using Cause as an Effect to construct a question: Why did the President ``probe into violence and deaths of people in wage-related protests''? Because the Commission was established -- does not answer the question.
    \item \textbf{Linguistic:} “to probe into the violence and the deaths of 44 people in wage-related protests” causes “The three-member Farlam Commission , chaired by retired judge Ian Farlam , was established by President Jacob Zuma”.
\end{enumerate}

\begin{enumerate}[label*=(\arabic*),resume*=example]
  \item{Chale was allegedly chased by a group of about 30 people and was hacked to death with pangas, axes and spears.}
\end{enumerate}
\begin{itemize}[leftmargin=1cm]
    \item \emph{Non-causal}: This sentence exhibits temporal events that have no causal connections. We highlight the tests it fails below:
\end{itemize}
\begin{enumerate}[leftmargin=1cm]
    \item \textbf{Why:} Why was Chale “hacked to death with pangas, axes and spears”? Because “Chale was allegedly chased by a group of about 30 people”. -- This test fails since the Cause does not really explain the reason for why there was the Effect.
    \item \textbf{Counterfactual:} The counterfactual is ``The hacking (Effect) is unlikely to occur if the chasing did not take place (NegCause).'' -- This test fails since hacking can occur without chasing. In fact, without chasing, Chale might have been hacked earlier.
\end{enumerate}

\begin{enumerate}[label*=(\arabic*),resume*=example]
  \item{``I observed the attack on the police, I have no doubt about it,'' Modiba said during cross-examination.}
\end{enumerate}
\begin{itemize}[leftmargin=1cm]
    \item \emph{Causal}: This sentence contains a description of an implicit causal event relation within a speech.
\end{itemize}
\begin{enumerate}[leftmargin=1cm]
    \item \textbf{Why:} Why did Modiba ``have no doubt about it''? Because Modiba ``observed the attack on the police''.
    \item \textbf{Temporal order:} Modiba must observe (Cause) before claiming he had no doubts (Effect).
    \item \textbf{Counterfactual:} Modiba cannot claim that he has no doubts (Effect) if he had not observed the attack (NegCause). 
    \item \textbf{Ontological asymmetry:} Using Cause as an Effect to construct a question: Why was it that Modiba ``observed the attack on the police''? Because he had no doubt about it -- does not answer the question.
    \item \textbf{Linguistic:} “I observed the attack on the police” causes “I have no doubt about it”.
\end{enumerate}

\begin{enumerate}[label*=(\arabic*),resume*=example]
  \item{Both Anoop and Ramanan are also accused in the case related to attack on the Nitta Gelatin office at Panampilly Nagar last year.\label{eg_6}}
\end{enumerate}
\begin{itemize}[leftmargin=1cm]
    \item \emph{Non-causal}: The inclusion of “also” in “also accused” renders that when constructing an answer for the Effect, we need to explain why they are accused in addition to someone else that has already been accused. Therefore, when performing the ``Why'' test, the answer seems incomplete. Hence, we do not consider this sentence to be causal.
\end{itemize}
\begin{enumerate}[leftmargin=1cm]
    \item \textbf{Why:} Why were ``both Anoop and Ramanan also accused in the case''? Because of the “attack on the Nitta Gelatin office at Panampilly Nagar last year .” -- This does not answer the question.
    \item \textbf{Linguistic:} “attack on the Nitta Gelatin office at Panampilly Nagar last year ” cause “Both Anoop and Ramanan are also accused in the case”. -- This does not work.
\end{enumerate}

\begin{enumerate}[label*=(\arabic*),resume*=example]
  \item{Striking mineworkers have threatened to halt all mining operations if their employers do not accede to their pay demand.
  \label{eg_7}}
\end{enumerate}
\begin{itemize}[leftmargin=1cm]
    \item \emph{Causal}: This sentence contains \textsc{Condition} relations, which we interpret as causal.
\end{itemize}
\begin{enumerate}[leftmargin=1cm]
    \item \textbf{Why:} Why could ``halt all mining operations'' occur? Because ``their employers do not accede to their pay demand''.
    \item \textbf{Temporal order:} Their pay demands must be acceded (Cause) before the halting of mining operations (Effect).
    \item \textbf{Counterfactual:} The halting of mining operations (Effect) is unlikely to occur if their pay demands were acceded (NegCause). 
    \item \textbf{Ontological asymmetry:} Using Cause as an Effect to construct a question: Why was it that ``their employers do not accede to their pay demands''? We cannot find an answer span for this question in the original Effect span.
    \item \textbf{Linguistic:} ``their employers do not accede to their pay demand'' causes ``halt all mining operations''.
\end{enumerate}

\begin{table*}[!ht]
\begin{center}
\begin{tabular}{llllll}\\\hline
Setting & F1 & P & R & Acc & MCC \\\hline
All \emph{Causal} & 72.28 & 56.59 & \textbf{100.00} & 56.59 & 0.00\\
Random & 55.72 & 56.61 & 54.92 & 50.66 & 0.00 \\
CNC Training & \textbf{78.22} & 72.68 & 84.66 & \bf 73.31 & \bf 45.15 \\
PDTB-3 & 56.68 & 66.41 & 49.43 & 57.23 & 16.90 \\
PDTB-3 Bal & 68.75 & 56.62 & 87.50 & 54.98 & 0.14 \\
CTB & 33.79 & \bf 86.05 & 21.02 & 53.38 & 23.80 \\
CTB Bal & 65.23 & 71.14 & 60.23 & 63.67 & 28.15\\
\hline
\end{tabular}
\caption{Metrics from predictions with the LSTM model on CNC Test Set using different training sets. F1, Precision (P), Recall (R), Accuracy (Acc) and Matthews Correlation Coefficient (MCC) are reported in percentages (\%). Highest score per result column is indicated in boldface.}
\label{tab:results-lstm}
 \end{center}
\end{table*}

\begin{enumerate}[label*=(\arabic*),resume*=example]
  \item{The strike will continue till our demands are conceded.
  \label{eg_8}}
\end{enumerate}
\begin{itemize}[leftmargin=1cm]
    \item \emph{Causal}: This sentence contains \textsc{Negative-Condition} relations, which we interpret as preventive causal type.
\end{itemize}
\begin{enumerate}[leftmargin=1cm]
    \item \textbf{Why:} Why could the event that ``the strike will continue'' be prevented? Because ``our demands are conceded''.
    \item \textbf{Temporal order:} The demands must be conceded (Cause) before preventing the strike from continuing (NegEffect).
    \item \textbf{Counterfactual:} The strike cannot be stopped from continuing (NegEffect) if the demands are not conceded (NegCause). 
    \item \textbf{Ontological asymmetry:} Using Cause as an Effect to construct a question: Why was it that ``our demands are conceded''? Because ``the strike could be stopped'' -- does not answer the question.
    \item \textbf{Linguistic:} “Our demands are conceded” prevents “the strike will continue”.
\end{enumerate}

\section{Other Baseline Models} \label{adx:other-baseline-models}

% \begin{table*}[!ht]
% \begin{center}
% \begin{tabular}{llllll}\\\hline
% Setting & F1 & P & R & Acc & MCC \\\hline
% All \emph{Causal} & 72.28 & 56.59 & \textbf{100.00} & 56.59 & 0.00\\
% Random & 55.72 & 56.61 & 54.92 & 50.66 & 0.00 \\
% CNC Training & \textbf{78.22} & 72.68 & 84.66 & \bf 73.31 & \bf 45.15 \\
% PDTB-3 & 56.68 & 66.41 & 49.43 & 57.23 & 16.90 \\
% PDTB-3 Bal & 68.75 & 56.62 & 87.50 & 54.98 & 0.14 \\
% CTB & 33.79 & \bf 86.05 & 21.02 & 53.38 & 23.80 \\
% CTB Bal & 65.23 & 71.14 & 60.23 & 63.67 & 28.15\\
% \hline
% \end{tabular}
% \caption{Metrics from predictions with the LSTM model on CNC Test Set using different training sets. F1, Precision (P), Recall (R), Accuracy (Acc) and Matthews Correlation Coefficient (MCC) are reported in percentages (\%). Highest score per result column is indicated in boldface.}
% \label{tab:results-lstm}
%  \end{center}
% \end{table*}

As described in Section \ref{sec:experiments}, in addition to the BERT model, we constructed a LSTM model as one of our baselines, considering its ability to account for long-distance relations between words. In this model, firstly, an embedding layer initialized with fastText’s Common Crawl 300-dimensional embeddings \cite{bojanowski2017enriching}. We experimented with GloVe 300-dimensional embeddings \cite{pennington2014glove} and a concatenation of GloVe and fastText in our initial experiments and selected fastText because it outperformed others. This layer is followed by two bi-directional LSTM layers with a dense layer on top. The final prediction is obtained via a softmax layer. For our training parameters, we used a maximum sequence length of 128, batch size of 32, a learning rate of $1e^{-3}$ with Adam optimizer and epochs of 20 with early stopping patience of 5. The obtained results are summarized in Table \ref{tab:results-lstm}. We observed that performance of this system is significantly lower than BERT for all scenarios we tested. However, conclusions regarding the transferability of PDTB-3 and CTB Bal onto CNC (based on F1 and MCC scores) holds.

\color{black}

\end{appendices}

\end{document}